# A Survey on Hough Transform, Theory, Techniques and Applications

**Allam Shehata Hassanein , Sherien Mohammad, Mohamed Sameer, and Mohammad Ehab Ragab**

**Informatics Department, Electronics Research Institute,**
**El-Dokki, Giza,12622, Egypt**

**Abstract**

For more than half a century, the Hough transform is ever-expanding for new frontiers. Thousands of research papers and numerous applications have evolved over the decades. Carrying out an all-inclusive survey is hardly possible and enormously space-demanding. What we care about here is emphasizing some of the most crucial milestones of the transform. We describe its variations elaborating on the basic ones such as the line and circle Hough transforms. The high demand for storage and computation time is clarified with different solution approaches. Since most uses of the transform take place on binary images, we have been concerned with the work done directly on gray or color images. The myriad applications of the standard transform and its variations have been classified highlighting the up-to-date and the unconventional ones. Due to its merits such as noise-immunity and expandability, the transform has an excellent history, and a bright future as well.

## 1. Introduction

In 1962 Paul Hough introduced an efficient method for detecting lines in binary images [1]. Spatially extended patterns are transformed to produce compact features in a parameter space. In this way, the Hough transform (HT) converts a global detection problem in the image space into an easier local peak detection problem in the parameter space. To describe the working of the HT algorithm, the slope-intercept parameterization and the voting scheme are summarized in Fig.1.

For the algorithm of Fig. 1, all edge pixels lying on a line cooperate to increase the content of the cell in the accumulator array corresponding to the real slope and intercept of that line. This shows strong evidence that the corresponding straight line really exists in the image. In this way, the HT can be viewed as a vote-counting-procedure where each edge point votes for all parameter combinations that could have produced it. For example, Fig. 2 depicts a line in the image space and the obtained accumulator array in the Hough space.

```
1. Build a parameter space with a suitable
   quantization level for line slope m
   and intercept c
2. Create an accumulator array A(m,c)
3. Set A(m,c) = 0  ∀ (m,c)
4. Extract image edges using Canny
   detector
5. For each pixel on the image edges
   (x_i, y_j)  ∀ (m_k, c_l) verifying equation:
   c_l = -x_i m_k + y_j
   Increment: A(m_k, c_l) = A(m_k, c_l) + 1
6. Find the local maxima in A(m,c) that
   indicate the lines in the parameter
   space.
```

Fig. 1 Algorithm of the HT (slope-intercept parameterization) [1], ∀ denotes "for all".

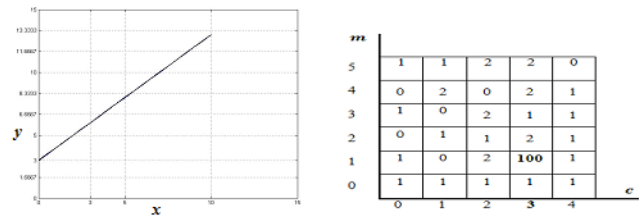

Fig. 2 A line in image space is represented as a cell: ($m$=1, and $c$ =3) in parameter (Hough) space.

So far, the slope intercept parameterization seems quite acceptable. However, we face a degenerate situation when having vertical lines (or even when the lines are close to vertical). In this case, the slope $m$ approaches infinity. This is the stimulus behind suggesting the ($\rho\theta$) parameterization in [2]. The parameter $\rho$ is the perpendicular distance from a chosen origin in the image space to the line, and the parameter $\theta$ is the angle between the perpendicular and the horizontal axis. In fact, this is a point-to-curve transformation using the normal representation of the line:

$$x \cos \theta + y \sin \theta = \rho \qquad (1)$$

As shown in Fig. 3, the multiple pairs ($\rho, \theta$) verifying equation (1) for a certain point: ($x_i, y_i$) is a sinusoidal curve. Each pair ($\rho, \theta$) corresponds to a candidate line passing through the point.

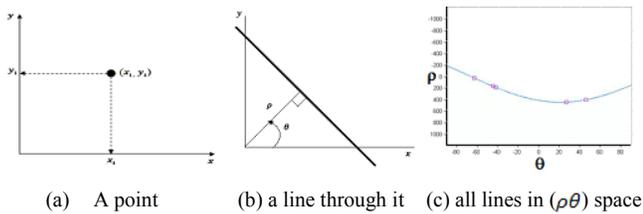

(a)   A point     (b) a line through it   (c) all lines in ($\rho\theta$) space

Fig. 3 A point-to-curve transformation in ($\rho\theta$) parameterization

The main advantages of using the HT are mentioned in the following lines. First, it treats each edge point independently; this means that the parallel processing of all points is possible which is suitable for real-time applications. Second, it can handle the cases of partially deformed and noisy shapes due to its voting scheme. Third, the HT can really detect multiple occurrences of lines since each occurrence has its specific cell in the parameter space. Moreover, as will be shown below, the HT can be extended to detect shapes other than lines. As a result of its several merits, the HT has a lot of applications such as, pipe and cable inspection, underwater tracking, road sign identification, lane detection, and many other industrial and medical applications. On the other hand, the HT has certain drawbacks. In particular, it generally requires a large storage and a high computational demand. For example, the number of calculations scales up rapidly with the dimensionality of the problem. However, the efficiency of the HT can be increased using prior knowledge to reduce the size of the accumulator array.

Comparative surveys of the HT are presented in [3], and [4]. However, they date back to more than a quarter century ago. The main contributions of our work are: explaining the most of HT variations elaborating on the basic ones, pursuing the solutions in literature for its drawbacks, underlining the transform direct application to gray and color images which maintain higher information content compared to the usually used binary images, and classifying its applications highlighting the up-to-date and unconventional ones.

The rest of this paper is organized as follows: the use of the HT in shape detection is shown in section 2, section 3 highlights the direct application of the transform to gray and color images, the speedup and memory saving approaches are explained in section 4, the applications are classified in section 5, and the survey is concluded in section 6.

## 2. Shape Detection

The concern of the standard Hough transform (SHT) is the identification of lines. However over the years, several variations have been suggested to identify other analytical or even irregular shapes. In this section, we will demonstrate the main aspects of using the HT in detecting the analytical and irregular shapes alike.

2.1 Line Detection

In this subsection, we elaborate on the use of the HT in detecting lines initially presented in the introductory section of this work. Due to their abundance, lines may be used in identifying many geometric entities [2]. Their use ranges from indoor industrial applications to outdoor vehicle surveillance. An example of the use of the HT in line detection is shown in Fig.4.

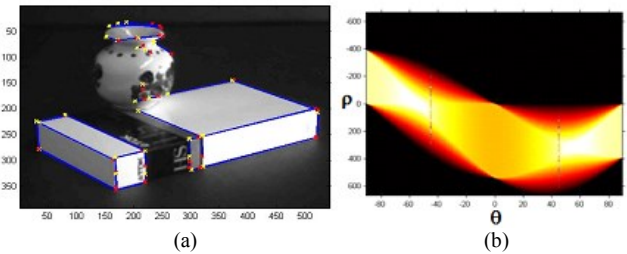

(a)                              (b)

Fig. 4 (a) Detected lines superimposed on image, (b) corresponding parameter space with intersection points representing lines.

Rectangular and Gaussian windows are two pre-processing methods presented in [5] for the high precision estimation of the HT line parameters. The two methods are compared to the SHT and have shown more accuracy with respect to real and synthetic images. A method for detecting corners as feature points is presented in [6] by using a 'patchy' HT where the corners are considered as intersections of straight lines. In [7], an approach is presented for using the HT in detecting line segments. This approach combines the dynamic combinatorial method (DCHT) proposed by Ben-Tzvi et al. in [8], and the dual plane HT method. The approach has bounds for its parameter space resulting in a low storage requirement, and an efficient implementation. In [9], a cascaded technique is shown to find line candidates efficiently. A modified algorithm is presented in [10] based on the likelihood principle of connectivity and thickness that makes short as well as thick line segments easier to detect in a noisy image. Some algorithms based on a probabilistic approach are proposed in [11], [12], and [13]. In contrast to conventional HT approaches, the probabilistic Hough transform (PHT) is independent of the size, shape, and arrangement of the accumulator array. The PHT has a lower bandwidth and a larger immunity against noise. The probabilistic and non-probabilistic approaches are compared in [14] where the randomized Hough transform (RHT) is introduced. Basically, $n$ pixels are randomly selected and mapped into one point of the parameter space instead of transforming each pixel into an $n-1$ dimensional hypersurface in the parameter space. The RHT versions included therein are dynamic RHT, random window RHT, window RHT, and the connective RHT. They are tested for line detection in synthetic and real images. It is shown in [14] that previous PHT approaches differ from the RHT in both randomizing and accumulating. In [15], a progressive probabilistic Hough Transform (PPHT) is utilized to minimize the computation. The PPHT uses the minimum fraction of votes needed to reliably detect the lines. This is based on the prior knowledge about the points composing the line.

The grid Hough transform (G-HT) is presented in [16] to detect the pitch lines in sporting videos. The G-HT is fast and memory-efficient since it uses the active gridding to replace the random point selection in the RHT. According to [16], forming the linelets from the actively selected points is more efficient than the random selection. Additionally, a statistical procedure is used to get rid of the expensive voting procedure. There are three merits for using the active gridding. First, we can locally evaluate whether the pairs on the same grid define linelets or not. Second, the isolated selected points will not be involved in the further computation being away of the linelets. Last, the number of linelets is a small fraction of the number of all the edge points. Hence, computing the measure function becomes much faster. The statistical Hough transform (S-HT) is estimated in [17] using histograms and kernel estimates. Due to its global nature, it is robust to noise and to the choice of the origin of the spatial coordinates, and all its parameters are automatically estimated.

A vanishing point detection algorithm is presented in [18] where the lines are detected. Then, using a coordinate transform, circles are translated into lines. Finally, the HT is used for detecting lines again. Thus, the positions of the vanishing points correspond to the line parameters in polar coordinates. Another vanishing point detecting approach is proposed in [19] being based on the k-means clustering. In this approach, there is no need to find the intersection of lines to extract the vanishing points. Therefore, both of the complexity and the processing time of the algorithm are reduced.

The work in [20] addresses a common misconception in the polar line parameterization when detecting the vanishing points. In the $\rho - \theta$ line parameterization, the vanishing points do not map onto lines. However, lines can be used as an approximation when the vanishing point is sufficiently distant from the image area. The error of this approximation is proportional to the ratio of the radius of image area to the distance of the vanishing point from the image center. The work also studies in detail the point-to-line-mappings (PLTM) as a class of line parameterizations having the vanishing points mapped onto lines. A mathematical background of the cascaded Hough transform (C-HT) is formulated providing the mathematical tools for constructing the PTLM triplets. This can be also used in detecting checkerboard structures utilizing the vanishing points. In this way, the work in [20] can be considered as an extension of that in [9].

In [21], the HT line detection is enhanced by compensating for the lens distortion. In this approach the search space is 3D, including the orientation, the distance to the origin and also the distortion. The algorithm presented in [22] is used for detecting 3D lines using the HT. This approach is applied to the range images of scenes containing boxes or blocks. Initially, peaks are found in the 2D slope parameter space. Then, peaks in the intercept parameter space are sampled. Line segments are fitted locally to the edge points detected by a Laplacian operator to accurately estimate the directions in the 3D.

2.2 Other Analytical Shapes

As mentioned above, the original use of the HT is detecting lines. However, it has evolved over the years to identify other analytical shapes such as circles and ellipses [23]. Both of the HT and the circular Hough transform (CHT) usuallly depend on converting gray-scale images to binary images. They both use a preliminary edge detection technique such as Sobel or Canny. Additionally, they both depend on a voting scheme. However, the CHT relies on the equation of circle having three parameters. This requires more computation time and memory storage, and increases the complexity. A basic algorithm of the CHT is shown in Fig. 5, while some examples of using it are demonstrated in Fig.6.

A comparative study of the CHT is introduced in [24]. The comparison considers the SHT for circles, the fast Hough transform (FHT), and the space saving approaches devised by Gerig and Klein [25]. A two-step algorithm [26] is presented for circle detection through a 2D HT and a radius histogram. The first step uses the HT for the detection of the centers of the circles and the second step validates their existence by radius histograms. The fine-tuning of the HT parameter space is presented in [27] using the genetic algorithms (GA). Additionally, a size invariant circle detection method is presented in [28] where a combination of variations of the CHT is suggested. The variations have an equivalent effect to applying a scale invariant kernel operator. They include adding edge directions, simultaneous consideration of a range of radii, using a complex accumulator array with a phase proportional to the log of radius, and a filter implementation.

```
1. Extract edges of the image e.g. using Canny
   detector.
2. Create a 3D Accumulator Array of zeros,
   A[x,y,r] for center locations and radius
   values according to image dimensions.
3. For each edge point, (xᵢ,yᵢ) in the image:
   3.1 For each candidate radius r:
      3.1.1 For each candidate x for center
            horizontal location:
         3.1.1.1 Find each possible y verifying
                 the equation:
                 (xᵢ – x)² + (yᵢ – y)² = r²
         3.1.1.2 Increment the cell:
                 A[x,y,r] = A[x,y,r] + 1
4. Search for local Maxima in A[x,y,r], to
   obtain the center locations and the radii
   of the circles in the image.
```

Fig. 5 A basic algorithm for the CHT

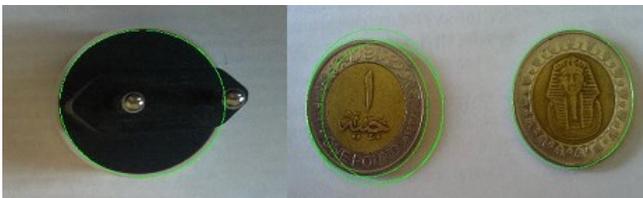

Fig. 6 Applying the CHT to three objects, the shades affect the detection of the coin in the middle.

A fast randomized HT (FRHT) for circle and circular arc recognition is presented in [29]. In this approach, one point was picked at random as a seed point, then a checking rule is used to confirm if it is truly on the circle or not. The FRHT requires less computation time and is more suitable for dense images compared to the previous RHT versions.

A sphere detection method based on the hierarchical HT (HHT) is introduced in [30]. The HHT has privileged features such as: reducing storage space and calculation time significantly, being robust with respect to noise and artifacts, and being applicable in 3D images. This technique is successfully applied in analysing the 3D structure of the human ball joints in the hip and shoulder which renders it useful in surgery planning. It can be used in detecting partially hidden and overlapping parts of a sphere. A method for great-circle detection on a sphere using the RHT is presented in [31]. It transforms central-projection images to spherical images mapped onto a unit sphere. In this way, great circles on the sphere correspond to lines on 3D planes easing their identification. In this work, the duality between a point and a great circle on a sphere has been exploited.

An approach of concentric circle detection is proposed in [32]. The image is pre-processed by performing edge detection, and then the circle centers are allocated using the gradient HT. Finally the radii are found using the one dimensional (1D)-HT. The detection efficiency is enhanced by the image discretization, and by reducing the resolution ratio in the process of circle center detection. The proposed combination of the gradient HT and the 1D-HT is reliable to noise and can deal with distortion, and edge discontinuity. An efficient RHT for circle detection is presented in [33]. It uses a probability sampling to optimize determining the sample points and assuming the candidate circles. Due to these optimizations, the sampling validity is improved and many false circles are filtered out. Circle detection is carried out in [8] using the DCHT to save in memory and computation time.

An extension is suggested in [34] to detect approximate circles and circular arcs of varying gray levels in an edge-enhanced image. Circles of arbitrary radii are detected in [35]. The algorithm performs the 3D convolution of the input image with the Hough kernel. The complexity is $O(n^3 \log n)$, which is better than $O(n^4)$ of the original CHT. A comparison between the CHT and the modified Canny edge detector for circle detection is carried out in [36] favouring the latter. According to [37], the main drawback of the CHT is the dependence on the result of the edge detector. Additionally, the CHT investigates a pre-defined radius value of the circle which may limit its use. Efforts are made in [35] to solve the radii definition problem by following a radius-invariant approach.

With respect to ellipse detection, an ellipse can be defined by five parameters, its center $(x_c, y_c)$, the major axis $a$, the minor axis $b$, and the slope $\theta$ as shown in Fig.7.

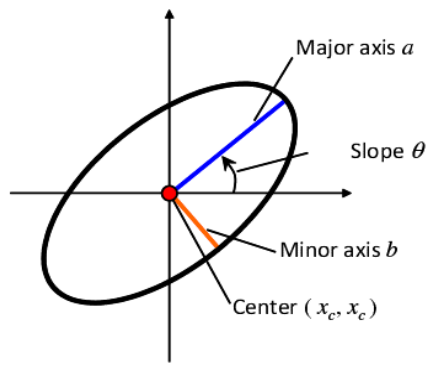

Fig. 7 Ellipse parameters.

Therefore, we need a five-dimensional parameter space for ellipse detection. If $N$ is the size of each dimension of the parameter space, an order of $O(N^5)$ is required to store the parameter space. Additionally, it takes time of order $O(N^5)$ to vote for each edge point and to search in the accumulator array [38] where a graphic processing unit (GPU) is used for speeding up. A method is presented in [39] for detecting ellipses by finding symmetric lines with the help of the linear Hough transform (LHT). In this method, global geometric symmetry is used to locate all possible centers of ellipses. Then, all feature points are classified into sub-images according to their corresponding centers. Such classifications save much time.

Curve detection methods based on the RHT are discussed in [40], and [41]. As mentioned above, the advantages of the RHT are a small storage, a high speed, and a flexibly extending parameter space. A real-time technique based on the RHT for ellipse detection is introduced in [42] to aid a robot in navigation and object detection. This technique uses clustering to classify all the detected ellipses into their predetermined categories in real time. The iterative RHT (IRHT) is developed in [43] for the detection of incomplete ellipses in the presence of high noise. It applies the RHT iteratively to a region of interest determined from the latest estimation of ellipse parameters.

With respect to other analytical shapes, an algebraic approach to the HT is discussed in [44] to lay a theoretic foundation for detecting general algebraic objects such as affine schemes. This approach is particularly useful in the problem of detecting special formations in medical and astronomical images. In [45], a detection method of polygons is suggested. This method finds image edges by utilizing the Canny algorithm. Then using the HT, the lines on each edge are drawn. Finally, the start and end points of the longest lines are found to detect the polygon present in the image.

2.3 Irregular Shape Detection

Various variations of the HT such as the SHT and the CHT are mainly used for detecting analytical curves using suitable parameterization. However, if the shape does not have an analytical form, we resort to using the generalized Hough transform (GHT) [46]. This transform uses a look-up table defining the relation between the boundary (positions and orientations), and the Hough parameters as a substitute for a parametric equation. Fig. 8 shows the GHT parameters, and Fig.9 displays its algorithm.

The GHT algorithm displayed here assumes that the sought model has the same scale and orientation as that of the look-up table. However, the rotations and scale changes can be dealt with using appropriate transformations. A remarkable attraction of the GHT is its ability to be applied hierarchically to build complex shapes from simpler ones. This makes the GHT a scalable transform.

The drawbacks of the high dimensional parameter space encountered by the analytic HTs can be avoided by the use of the GHT [46]. The advantages of the GHT include being robust to occlusion and the presence of additional structures (e.g. other lines, curves, etc.), being tolerant to noise, and being able to find multiple occurrences of a shape during one pass. The main problems of it are the high storage and computation demands. Some improvements are suggested for the modified version of the GHT in [47]. It uses a pair of boundary points with the same slope angle to derive rotation-invariant parameters. Additionally, the speed is enhanced by applying the FHT. Each vote is contributed by a pair of edge pixels with the same slope angle. In this method, the 4D Hough domain is trimmed down to 2D eliminating the false votes. A comparative study of the GHT techniques is shown in [48]. It investigates the performance of some efficient GHT techniques e.g. an extension of the Thomas's rotation-invariant algorithm [49]. An approach to shape matching for image retrieval is presented in [50]. This approach makes use of the GHT and works well in detecting arbitrary shapes even in the presence of gaps and under rotation, scale and shift variations. It also uses a preliminary automatic selection of the appropriate contour points to reduce the computation.

A probabilistic framework is shown in [51] for object detection. It is a variation of the HT that has similar simplicity and wide applicability. However, it bypasses the problem of multiple peak identification in the Hough space and permits multiple object detection without invoking non-maximum suppression heuristics. This framework can reuse the training procedures and the vote representations developed for implicit shape models or Hough forests. That boosts its use in many applications such as: cell detection and counting pedestrians. Nevertheless, the computation time sometimes increases compared to the traditional HT. An approach of 3D image processing based on the modified GHT is shown in [52]. The indexing of the look-up table is modified seeking for a more efficient performance.

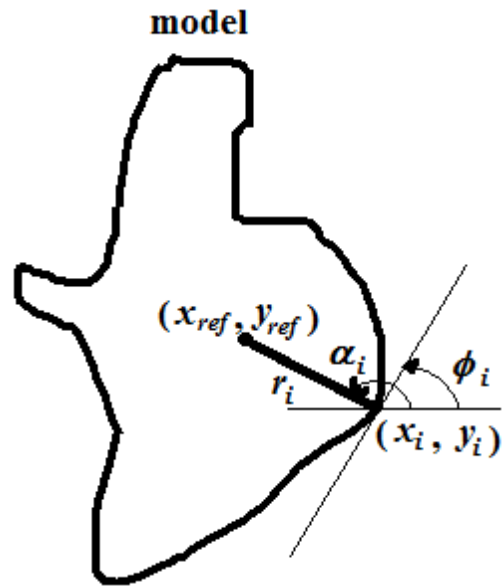

Fig. 8 Parameters of the GHT

```
Constructing Look-up Table of Model (usually once and off-line)
    1. Extract edges of the model image e.g. using Canny detector.
    2. Choose any point inside edges as a reference, (x_ref, y_ref).
    3. For each edge point , (x_i, y_i):
        3.1    Compute angle between x-axis and slope direction of contour, φ_i, and displacement
               vector to reference, r_i.
        3.2    Quantize φ_i according to required precision.
        3.3    Compute distance to reference, ‖r_i‖, and angle α_i between x-axis and r_i.
        3.4    Store ‖r_i‖, and α_i in table in the row of quantized φ_i.
Searching for Model in New Images (assuming same model scale and orientation, equivalent to searching
for the reference point)
    1. Extract edges of the image e.g. using Canny detector.
    2. Create Accumulator Array, A[x,y] for possible reference points
    3.  For each edge point , (x_i, y_i):
        3.1    Compute φ_i, quantize, and get all pairs (α_i, ‖r_i‖) from its row in look-up table.
        3.2    For each pair (α_i, ‖r_i‖), compute reference candidates:
```

$$x_{cand} = x_i + \|r_i\|\cos(\alpha_i) \quad \text{and} \quad y_{cand} = y_i + \|r_i\|\sin(\alpha_i)$$

```
        3.3    Increase candidate votes:
```

$$A[x_{cand}, y_{cand}] = A[x_{cand}, y_{cand}] + 1$$

```
    4. For local Maxima in A[x,y], (x_ref = x, y_ref = y)
```

Fig. 9 A basic algorithm for the GHT

## 3. Tone and Color

Most implementations of the HT require the pre-processing steps of binary-conversion and edge detection before being applied to gray images causing a loss of useful information. In this section, we demonstrate other approaches working on images with higher information content. They range from simply adding a gray value to the accumulator array to color segmentation and dealing with color histograms. The section is concluded with a case study of a gradient approach working directly on gray images for line detection. Additionally, this approach is compared to the conventional SHT.

### 3.1 Irregular Shape Detection

An extension of the HT, the gray scale Hough transform (GSHT) is presented in [53]. It aims at detecting thick lines (bands) in a gray scale image directly to save binary-conversion time and to have more information content. The GSHT enlarges the accumulator array by adding the gray value $G_i$ of each edge point as an additional dimension along with $\rho$ and $\theta$ (in contrast with the conventional HT). In this way, there is a high storage demand. A gray scale inverse HT is proposed in [54] to detect line segments. This is combined with the modified GSHT proposed in [53]. The method is capable of determining lines exactly as they appear in the original image pixel by pixel with its correct gray value. On the other hand, a technique is presented in [55] to extract homogeneous line segments in gray scale images without obtaining edges. A region has been defined as the union of several connected line segments and a line segment is defined as a straight line with the property of uniformity among the pixels on the line. The property of uniformity is the value of variance of gray values along the line. Accordingly, the line segment is defined using two threshold parameters (the minimum length of line and the maximum variation of line). Since multiple line segments can collaborate to define a region, the approach is capable of detecting regions irrespective of their shape and size.

The GSHT is extended to region extraction in color images in [56]. It depends on the homogeneity of color obtained from the trace of the covariance matrix of the pixels along lines. The homogeneity of color pixels on a line segment may be viewed in two ways. The first is component-wise (a set of pixels has to be homogeneous in each of the three components: red, green, and blue). The second is provided by combining the values of the set of pixels without looking at each one individually. However, the approach has some problems to tackle. One of them is that it may fail in detecting an object having many small uniform regions. Another problem is the critical choice of suitable quantization levels in the Hough space.

As mentioned in [56], there are three possibilities for detecting edges in color images. First, to detect them separately in individual spectral components (such as the red component) using local gradient operators. The resultant edge magnitude may be taken as, either the maximum edge value for all components or a linear combination of them. Second, to use the brightness difference or the ratio of the same pixel in different color components (which is a very informative feature). Third, to create a multi-spectral edge detector of size $2 \times 2 \times n$, as each pixel is represented by an $n$-dimensional vector. The $2 \times 2$ neighborhood is similar to that of the Roberts gradient operator.

The GHT with color similarity between homogeneous segments is presented in [57] for object detection. The use of color has made the algorithm robust enough to recognize objects in spite of being translated, rotated, scaled and even located in a complex environment. In [58], a spatial color histogram is proposed for segmentation and object tracking. This model encodes the color distribution and spatial information using the GHT as a location estimator.

3.2 Case Study: the Gradient Hough Transform

A line detection algorithm which works directly on gray images is the gradient Hough transform (GrdHT) introduced in [59]. The algorithm calculates the gradient magnitudes for the image pixels. Those with a gradient magnitude less than a certain threshold practically belong to constant intensity regions in the image. Such regions are ignored in the voting of the GrdHT since they are not likely to contain edges or lines. Additionally, the GrdHT has a sensitivity parameter which tunes its operation in a fuzzy manner. Besides working directly on gray images, the GrdHT has other advantages such as: working fast since the code does not have any loops, and being able to detect the ending points of a line segment.

We have carried out two sets of experiments. The first compares the GrdHT to the SHT, while the second probes the effect of varying the sensitivity parameter on the number of detected lines by the GrdHT.

In Fig. 10, the left column belongs to the GrdHT (with default sensitivity value), while the right belongs to the SHT. For the image in the first row, the GrdHT has a good performance in detecting most lines of the checkerboard pattern. While, the SHT seems to be more affected by the projective distortion. It only detects some of the horizontal lines parallel to the image plane. For the image in the second row, the performance of the GrdHT is worse since it is based on the derivative effect which amplifies the already high content of noise in this image. For the image in the third row, the GrdHT detects most of the lines in the image while the SHT misses the most. The reason for this is that when the image is converted to binary in the case of the SHT, it loses most of its information content. A poor performance is seen for the image in fourth row for both algorithms. The GrdHT overestimates the number of detected lines due to the rich texture and high noise content of the image. While the SHT underestimates this number especially for the horizontal beams of the railroad. The reasons for this are: most of them are short, and their sizes are affected by the projective distortion. In fact they disappear as they approach the vanishing point. Therefore, the votes obtained by each are few compared to that of the rails and not remarkably high compared with other small edges in the image.

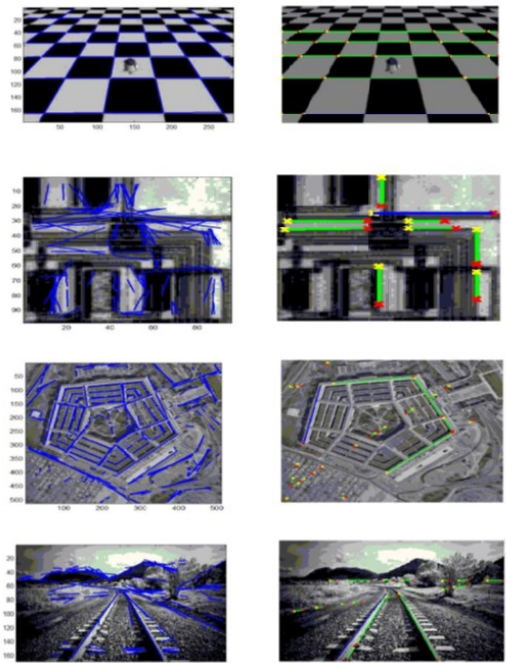

Fig. 10 Comparison between the GrdHT (left column), and the SHT(right column). Original images obtained from [59], and [60].

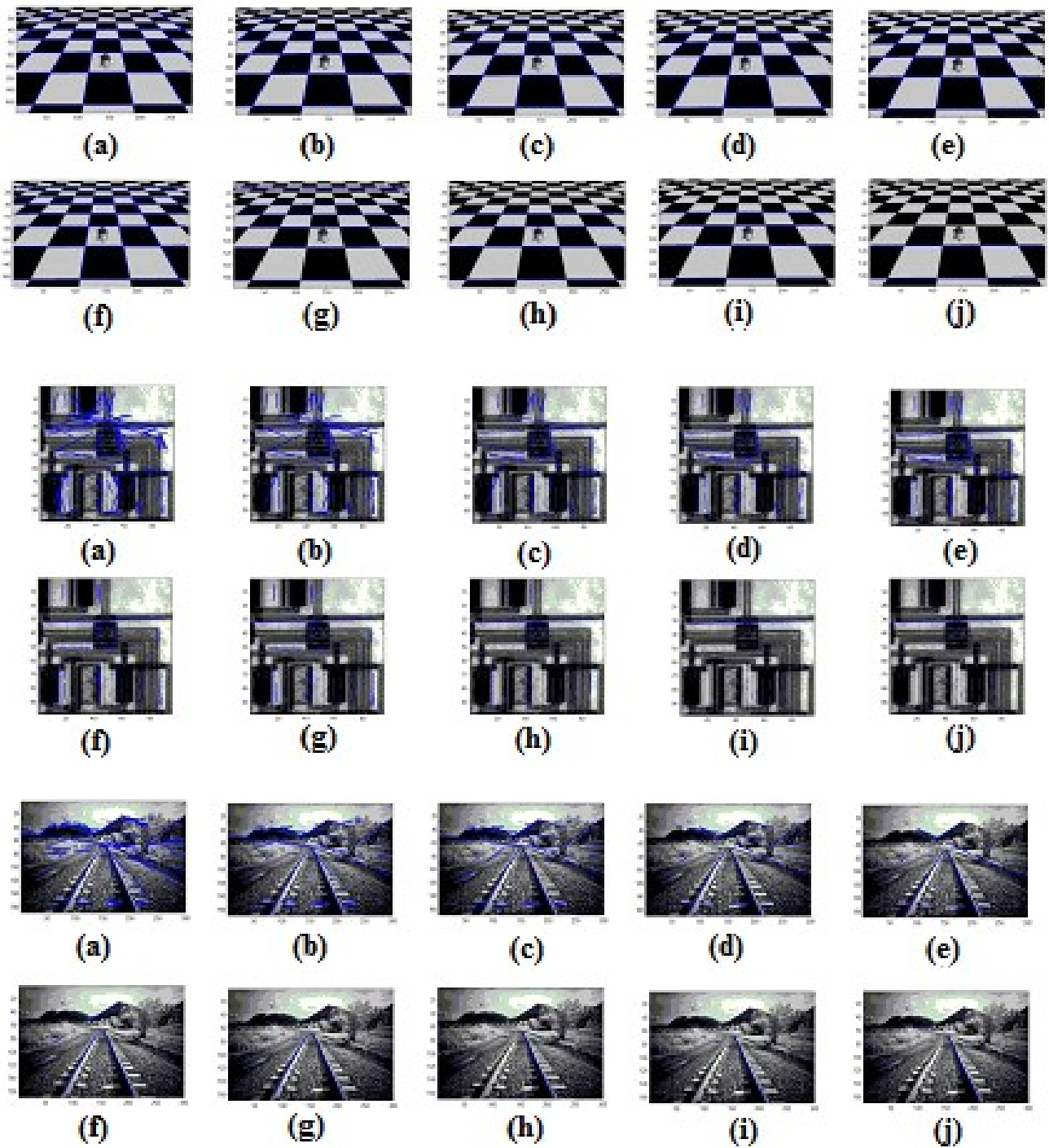

Fig. 11 Effect of varying the sensitivity parameter in the GrdHT on the number of detected lines (a) 0.02, (b) 0.04, (c) 0.06, (d) 0.08, (e) 0.1, (f) 0.2, (g) 0.3, (h) 0.4, (i) 0.5, (j) 0.6. Original images obtained from [59], and [60].

In Fig. 11, the effect of the sensitivity parameter is obvious for the all three images. The smaller its value, the more the lines that are detected. However, it is not easy to set the value of such parameter since its optimum value varies from an image to another.

To conclude this section, the GrdHT is more suitable for images which are affected by the projective distortion, and for images which lose most of their information content when converted to binary. The SHT is more suitable for images suffering from a high noise effect. Although the

GrdHT is more accurate than the SHT in many cases, setting its sensitivity parameter is a critical issue and is image-dependent.

## 4. Speedup and Memory Saving

While the HT possesses the mentioned benefits, it requires a long computation time and large data storage. A lot of research efforts have been made to ease such burdens. These efforts range from voting acceleration, to hardware implementations passing by parallel computing, and memory reduction. Some important aspects of these issues are shown below.

The field programmable gate array (FPGA) is among the most common hardware accelerators used in real-time implementations of the HT. For example in [61], the FPGA is used to store the Hough parameters in its memory block. This approach is not only memory efficient but can also be used for both line and circle detection. The hardware resources are minimized in an implementation of the HT shown in [62] due to exploiting both the angle-level and pixel-level parallelism. The run-length encoding is used to accelerate the HT by skipping the unnecessary computations and memory accesses. The real-time line detection in high-definition video combines the parallel Hough transform (P-HT) and the FPGA in an architecture-associated framework. The FPGA is used to build two modules. The first is a resource-optimized Canny edge detector with enhanced non-maximum suppression used to extract the candidate edge pixels for subsequent accelerated computations. The second is a multi-level pipelined P-HT for line detection. The use of the FPGA with the CHT is shown in [63]. This approach has a one-dimensional accumulator array stored in the FPGA memory and a 2D accumulator stored outside it.

To save in the computation time of the CHT, two paralleled methods are considered in [64]. The first is based on the threading building blocks (TBB) which is a set of C++ templates of parallel programming capabilities. The second uses the compute unified device architecture (CUDA) parallel computing platform on a GPU. The Hough voting used in the construction of the accumulator array is suitable for the parallel processing performed by both methods. According to the experimental results therein, the maximum speedup achieved on a 4-core system is 3.648x, while the maximum on the GPU platform is 45.7x. Another parallel implementation of the CHT is shown in [65]. Two methods for parallelization have been suggested. Each of which has been implemented on four different graphic cards using the CUDA. The proposed methods have been executed on GPUs verifying up to 65 times more speedup compared to the sequential algorithm.

The voting acceleration is employed in [11] where the PPHT is used for line detection. This approach minimizes the computation by using only a fraction of points for voting. On the other hand, an improved HT method is applied in [66] for detecting line segments in the presence of background clutter. A local operator in the parameter space is defined to enhance the discrimination between the accumulator peaks caused by real line segments and that result from noise. Meanwhile, a global threshold is obtained by analyzing both of the original accumulator histogram and the noise. The fuzzy generalized Hough transform (FGHT) is utilized in [67] for the retrieval of objects (instead of the conventional GHT) to alleviate the computation and memory requirements needed to handle the rotation and scale variations. For multiple object detection (five lumbar vertebrae), the work in [68] proposes combining the GA with the HT to reduce the search space complexity. Additionally, a probabilistic framework is proposed in [51] to solve the problem of multiple peak identification in the parameter space using the greedy inference.

Hough forest trees [69] are a combination of random forests and the GHT which is useful in real-time object detection and tracking. Human action classification with the help of the HT voting framework is shown in [70] and [71]. Random trees are exploited in the training phase to learn the mapping between densely-sampled feature patches and their corresponding votes in the Hough space. The leaves of the trees form a discriminative multi-class codebook in a probabilistic manner.

An effort has been made in [72] to optimize the HT performance. A centroid-based HT is used for the skew correction of the license plate recognition. The computation is reduced by focusing only on the centroid point of each character. Then, the HT is used to locate the line passing through these points and to find its skew. In [73], a crop row detection is proposed using a gradient-based RHT. It is found that by changing the mapping method, the gradient-based RHT reduces the computation burden meeting the real-time requirements. Memory saving is discussed in [48] where several GHT techniques are analyzed. In [52], an approach of 3D image processing modifies the indexing of the look-up table of the GHT. The total complexity of the algorithm decreases to a quadratic dependence on the number of samples defined by the object.

An eliminating particle swarm optimization (EPSO) algorithm is shown in [74]. The aim of EPSO is improving the speed having its parameters as the particle positions. Additionally, the work therein compares the EPSOHT, IRHT, and RHT in presence of noise. The compared approaches are close to each other with respect to accuracy and speed under low noise levels. However, as the quality of images worsens, the EPSOHT achieves the highest accuracy in the lowest time.

Some other approaches used to improve the HT performance are mentioned in this work. For example,

in [75] different hierarchy levels, and line partitioning into different blocks are used. Image compression is utilized in [76]. Speed is improved in [77] using the FHT. Additionally, end points are used instead of whole lines, and the gradient constraints are employed. Captured images are split in [78], and the HT is combined with a contour detecting algorithm in [79].

In this section, we have followed an inclusive approach. However, the crucial issues of improving speed and saving memory recur throughout the whole of this paper.

## 5. Hough Transform Applications

Due to the merits of the HT mentioned above e.g. noise immunity, it has been widely used in many applications. For example, 3D applications, detection of objects and shapes, lane and road sign recognition, industrial and medical applications, pipe and cable inspection, and underwater tracking. Some of these applications are illustrated below.

5.1 Traffic and Transport Applications

The hierarchical additive Hough transform (HAHT) is proposed in [75] to detect lane lines. The suggested HAHT adds the votes at different hierarchy levels. Additionally, line partitioning into different blocks reduces the computation burden. An approach of lane detection is suggested in [76], where the HT is combined with the joint photographic experts group (JPEG) compression. However, the algorithm is tested only in simulations. On the other hand in [77], the FHT is integrated by the gradient constraints for lane detection. It introduces a reverse voting strategy and gets the end points of the extracted line in order to represent it only by two points. In this way, the lane can be detected by matching two points instead of two parallel lines in the parameter space. Not only the combination of the FHT and the gradient improves the speed, but also increases the stability of the line extraction.

An adaptive unsupervised classifier is used in [80] to detect lanes. This is particularly useful for road estimation under changing system parameters. In this way, the system assists the driver to follow the road center. The inverse perspective map (IPM) is mad use of in [81] to get a top view of the road. Then, the HT is used to detect the lanes and to choose the most suitable for the vehicle to travel in. The IPM transformation is obtained with the help of information about the camera calibration and the pose assuming a flat road. The lanes appear in the IPM image as vertical and almost parallel straight lines. A lane departure warning system is proposed in [82] exploiting an optimized architecture of HTs. The suggested architecture has the ability of processing 16 LHTs simultaneously. The lane boundaries are detected and made available in [83] through a vehicular network to enhance the safety level and to boost the drivers' cooperation. The PPHT is used to extract the roads in [84] using only a small number of randomly selected edge points. This helps in the reduction of computation time. Utilizing high-resolution remote sensing images, the roads are extracted in [85] with the aid of the wavelet transform and the HT. The wavelet transform is used at the beginning to de-noise the image and to detect the road edges. Then, the HT is applied to enhance the road extraction. In [86], roads are extracted from high resolution synthetic aperture radar (SAR) images. The roads appear in the SAR images as homogeneous dark areas bounded by two parallel lines (or arcs). An average HT accumulates the gray level intensities of pixels on each line, and then divides the sum by the number of pixels. A system for driver assistance and lane marking detection and tracking is suggested in [78]. In this system, a real-time captured image is split. Then, the HT is applied to detect the lane boundaries. The reason for using splits is to alleviate the HT computation burden. After detecting the lanes in different splits, the partitions are merged to find the final most probable lane boundaries. A vision-based driver assistance system is proposed in [87] to promote the driver's safety in the night time. The suggested system carries out both of lane detection and vehicle recognition. Additionally, systems for detecting the driver-drowsiness can be built using the CHT to locate the eyes of the driver using a small camera mounted inside the car. In this case, the camera watches the face and observes if the eyes are closed for a number of consecutive frames.

A system for vehicle license plate (VLP) detection is shown in [79]. The system combines the HT and a contour detecting algorithm to improve the speed. Although the input images can be taken from various distances with inclined angles, the system is able to detect multiple VLPs. In [72], a perspective centroid-based HT for Skew correction of license plate recognition is shown. It gets first the centroid of each character on the plate, and then the HT is used to estimate a line passing by the centroids. Knowing the skew angle, the plate position can be corrected for a more accurate recognition.

An algorithm for the Chinese road sign recognition is proposed in [88]. It uses an improved two-step CHT to separately locate the center of the sign circle, and then to determine its radius. Additionally, a combination of the SHT and the GHT is used in [89] for road sign recognition. A project for rail road defect inspection is proposed in [90]. It finds the defects of the railroad tracks with a notification to the inspector of the type and location of the defect. For example, the CHT is used to detect missing bolts, defects, and cracks in ties.

## 5.2 Biometrics and Man-Machine Interaction

A generic hand detection model with articulated degrees of freedom is presented in [91]. It uses the geometric feature correspondences of points and lines. The lines are detected by the PHT. Then, they are matched to the model. The GHT is employed in [92] for the recognition of the Turkish finger spelling. Interest regions are obtained using the scale invariant features transform (SIFT). The irrelevant interest regions are eliminated with the help of the skin color reduction.
Additionally, a real-time human-robot interaction system based on hand gestures and tracking is proposed in [93]. It combines the connected component labeling (CCL), and the HT. Detecting the skin color, it has the ability to extract the center of the hand, the directions and the fingertip positions of all outstretched fingers. On the other hand, the HT is applied to finger print matching in [94]. The algorithm uses a robust alignment method (descriptor-based HT). Additionally, it measures the similarity between fingerprints by considering both minutiae and orientation field information.
The CHT is used to locate the eyes and the heads for the people detection in [95]. Additionally, the pupil is located in [96] using a combination of the Canny edge detector and the CHT. Moreover an iris localization algorithm is suggested in [97] where the CHT and the image statistics are used to estimate a coarse iris location in the images of the eye. The pupil is located using a bi-valued adaptive threshold and the two-dimensional shape properties. The limbic boundary is located by using the Hough accumulator and the image statistics. Finally, these boundaries are regularized using a technique based on the Fourier series and the radial gradients. Additionally, an iris segmentation method using the CHT and active contours is proposed in [98]. Since richer iris textures are closer to the pupil than to the sclera, the inner boundary of the iris is accurately segmented in its real shape, while the outer boundary is circularly approximated. In an eye image, the outer iris boundary is firstly obtained by the CHT. The pupil region of interest is defined accordingly. Then, the CHT is applied again to obtain an initial contour of the pupil for active contour segmentation. Such a close approximation, decreases the space search, and saves in the processing time. Furthermore, the CHT is used for person head detection, in [99].

## 5.3 3D Applications

In [22], a detector is devised for 3D lines, boxes, and blocks. Using 2D Hough space for parallel line detection, this method reduces the computational cost. The 3D HT is used for the extraction of planar faces from the irregularly distributed point clouds. The work also suggests two reconstruction strategies. The first tries to detect the intersection lines and the height of the rising edges. The second assumes that all detected planar faces should model some part of the building. The experiments show that the latter is able to reconstruct more buildings with more details, but sometimes generates artificial parts which do not exist in reality. To recognize objects in 3D, [100] proposes a 3D Hough space voting approach for determining object categories learned from artificial 3D models. The proposed method uses the ray voting for object reference points. This allows a cluster of votes to be shown in similar directions. The model can be trained with large amounts of data taken from different sources at varied scales. A modified GHT for the 3D image processing is presented in [52]. The principal modification in the GHT appears in the lookup-table indexing.
The airborne laser scanner technique is described in [101] as the most appropriate way to acquire 3D city models. Once the 3D lidar data are available, the next task is the reconstruction of building models. Specifically, the approach focuses on the detection of the roof planes. Among the approaches of 3D roof plane detection are the HT, and the random sample consensus (RANSAC). A comparison of both algorithms, in terms of the processing time and the sensitivity to the point distribution within the 3D cloud are shown therein. The RANSAC algorithm is faster and more efficient. Additionally, the HT is more sensitive to the segmentation parameter values. However, the RANSAC major limitation is that it sometimes generates 3D constructions that do not exist in reality. A proposed extension allows harmonizing the algorithm with the geometry of roofs. Once the roof planes are successfully detected, the automatic building reconstruction can be carried out. Additionally, the detection of vanishing points mentioned above in [9], [18], [19], and [20] is of crucial value for the 3D reconstruction.

## 5.4 Object Recognition

An approach is presented in [57] for object detection using the GHT with the color similarity between homogeneous segments of the object. The input of the algorithm is previously segmented regions with homogeneous color. According to this work, the approach is robust to illumination changes, occlusion and distortion of the segmentation output. It is able to recognize objects in spite of being translated, rotated, scaled and even located in a complex environment. The work shown in [102] proposes a robust method for recognizing objects among clutter and in the presence of occlusion. A close to real-time performance is achieved. The recognition of the objects depends mainly on matching individual features stored in a database of known objects using a fast nearest neighbor algorithm. The HT is then used to identify clusters belonging to a single object. In addition to this, recognition of multiple instances of an object is proposed in [103]. An active binocular vision system for localizing multiple instances of objects of the same-class in different settings is considered. The SIFT is used to find up to six object instances of the same class simultaneously. A probabilistic framework for multiple object detection is proposed in [51]. This work tries to solve the problem of multiple peak identification in the Hough space by applying the greedy inference which picks the overall (global) maxima. Additionally, [69] proposes a method using random forests to improve the GHT performance. The detection corresponds to the maxima of the Hough space about the possible location of the object centroid. Moreover a method for detecting moving objects is shown in [104]. An unsupervised moving object detection algorithm with an on-line GHT starts with simple motion estimation, and then selects the optimal foreground blobs based on multiple instance learning.

In [105], an attempt is made to broaden the scope of the HT to deal with more complicated patterns compared with primitive lines and circles. A method to extract convex hulls and other related features from the transform space is shown. In order to reduce the computation cost, two more efficient algorithms are proposed. The first is the piecewise-linear Hough transform (PLHT), and the second is the fast incremental Hough transform (FIHT). Furthermore, the HT is extended in a hierarchical form used with feature points.

In [106], an eight-neighborhood-based generalized HT (EN-GHT) for the recognition of multi-break patterns is proposed. The EN-GHT differs from the GHT in the configuration of the lookup-table which becomes more suitable for pattern recognition. Additionally, it is difficult for the conventional GHT to detect such broken areas. On the other hand, [67] proposes an approach of shape matching for image retrieval. The system selects the dominant points that can represent the shape for matching. Then, the voting is performed on the level of a region not a single pixel in the accumulator array. It is shown that the system works well in detecting arbitrary shapes even in the presence of gaps, rotations, different scales, and shift variations.

## 5.5 Object Tracking

In [107], a shape based tracking is performed using a mixture of uniform and Gaussian Hough (MOUGH). This approach is able to locate and track objects even against complex backgrounds such as dense foliage while the camera is moving. In [108], an object tracking in a sequence of sparse range images is suggested using the bounded Hough transform (BHT). This is a variation of the GHT which exploits the coherence across image frames. A method for rectangular object tracking is suggested in [109]. First, it finds the edges using the Canny edge detector, then applies the HT to find all lines in a predefined window surrounding the tracked object. A spatial color histogram is proposed in [58] for segmentation and object tracking. The spatial color histogram model encodes the color distribution and the spatial information. The GHT is used as a location estimator, which estimates the object location from a frame to another using its voting capabilities.

Planner pose estimation is tackled in [110]. The approach relates the 2D Hough space to the 3D information about the poses of planar objects. The HT is used to detect rectilinear segments grouped into a bounded figure in the planar surface. Then, a numerical representation of the Hough space is obtained. Comparing the numerical representations to that of the test patterns, the object pose is estimated. An indoor method for robot motion estimation is described in [111]. A line feature is detected and tracked in the robot fixed frame depending on the characteristics of the robot motion. The line tracking can be combined with the robot navigation. The HT is used to match each tracked line to its corresponding wall.

The HT is utilized for the automatic track initiation of moving objects in [112]. It is used in the radar surveillance applications. The reasons for this are having many objects for the sensor to observe, and the received measurements usually are noisy with uncertain origin. This makes the HT noise immunity highly advantageous in such settings. Furthermore, a multiple track initiation method using the HT is proposed in [113] for targets moving with constant acceleration in a heavy clutter environment. The SHT is used to filter the clutter. Then, using the RHT and the minimum distance metric, the candidate tracks can be followed. Finally, the actual track is found by subtractive clustering.

## 5.6 Underwater Applications

A method for geometrical shape recognition for an underwater robot images is proposed in [114]. The recognition problem is transformed into a bounded error estimation problem compared to the classical GHT. An underwater system is developed in [115] to carry out visually guided tasks on an autonomous underwater vehicle (AUV). Underwater pipelines are detected using the HT. Then, the binary morphology is employed for determining their orientations. In [116], the HT and an AUV are used for cable tracking. Additionally, in [117], a micro remotely operated vehicle (micro-ROV) is used to test the net integrity in fish cages by analyzing the frames of a video camera exploiting the PPHT.

In [118], the PHT is used to locate underwater pipelines. It searches for line segments among the edges and returns many line segments per pipe outline. The Hough pruning algorithm merges the semi-collinear lines. In [119], a model is suggested to detect and track underwater pipelines in complex marine environments. The images are filtered to reduce noise and to be enhanced. Then, the HT is used to get the boundary of the pipelines parameterized as curves. The curves with extracted features are used for estimating the locations and orientations of the pipelines. A marine tracking system is proposed in [120] where the automatic detection and tracking of man-made objects in subsea environments are pursued under the circumstances of poor visibility and marine snow. The proposed system involves minimizing the noise and video artifacts, estimating the camera motion, detecting the line segments and tracking the targets.

## 5.7 Industrial and Commercial Applications

There are myriad industrial and commercial applications of the HT and its many variants. However, for the sake of focusing on the trends, only a few has been mentioned in this subsection. A knowledge-based power line detection method is proposed in [121] using an unmanned aerial vehicle (UAV) surveillance and inspection system. A Pulse coupled neural network (PCNN) filter is developed to remove the background noise from the image frames before using the LHT. The knowledge-based line clustering is applied to refine the results. A variation of the HT which decides whether an item under inspection passes or fails is presented in [122]. A visual inspection system for industrial defect detection is presented in [123]. Where, the HT is used to check clamps, holes, and welds. The CHT is used in [124] for pipe inspection and to build a model for the defect. The detection of pipe joints is achieved by looking for peaks in the Hough space using the circle parameters of the pipe joint which is represented as a circle.

In [125] a prototype for cell detection algorithm using the CHT is proposed. Although the cells in a bee comb have a hexagonal shape, the spaces in the cells are approximated as circles. That makes it possible for the application to count the number of cells in a bee comb. Both of the CHT and the LHT have been combined in [126] for detecting the botanical and mineralogical alterations that result from the natural seepage of carbon dioxide and light hydrocarbons. The CHT is used to distinguish the holes from the background then collinear centers of the detected circles are found using the LHT.

Additionally, [127] proposes a skew detection for the barcode images. In this application, the HT is used to detect the scanned lines of the barcode. Then, a correction module is used to rotate the barcode to the upright position.

## 5.8 Medical Applications

The medical examination approach in [128] is based on a warped frequency transform (WFT) to compensate for the dispersive behavior of the ultrasonic guided waves, followed by a Wigner-Ville time-frequency analysis and the HT to further improve the defect localization accuracy. In [68], the positions and borders of vertebrae are automatically detected. The HT is combined with the GA to determine the moving vertebrae, concurrently.

A real-time tracking system of surgical instruments in laparoscopic operations is proposed in [129]. This system combines the particle filter tracking with the HT. Due to the condensation algorithm of the particle filter, the approach performs well in heavy clutter. Additionally, the HT is robust under illumination changes, occlusion and distractions. The Hough accumulator is constructed using the gradient direction image calculated by the principal components analysis (PCA). This improves the accuracy of the obtained edge orientation and speeds up the computation of the HT.

## 5.9 Unconventional Applications

To emphasize the versatility and the ever-expanding nature of the HT, we end this section by some unconventional uses. A data mining approach is suggested in [130] to find the arbitrarily oriented subspace clusters. This is accomplished by mapping the data space to parameter space defining the set of possible arbitrarily oriented subspaces. The clustering algorithm is based on finding the clusters that accommodate many database objects. The approach can find subspace clusters of different dimensionality even if they are sparse or intersected by other clusters within a noisy environment.

An algorithm for feature matching is presented in [131]. The feature matching problem is cast as a density estimation problem in the Hough space spanned by the homography hypotheses. Specifically, all the correspondences are projected into the Hough space, and matches are determined. It deserves mention that the SIFT algorithm has a HT step incorporated within [102]. Moving from features to a higher scale, a classification of a scene based on its structure and the set of lines therein is described in [132].

# 6. Conclusions

The Hough transform has attracted a lot of research efforts over the decades. The main motivations behind such interest are the noise immunity, the ability to deal with occlusion, and the expandability of the transform. Many variations of it have evolved. They cover a whole spectrum of shape detection from lines to irregular shapes. New variations are expected to appear moving the transform closer towards the recognition of more complex objects. Mostly, the transform and its variants have been applied to binary images. However, this is changing; we have shown some work done directly on gray and color images. We expect that more work will be done on color images maintaining most of the information content. This will be promoted with the persistent research efforts done for more memory-efficient and speedy implementations. Nowadays, numerous applications have made use of the Hough transform in many fields such as traffic, biometrics, object recognition and tracking, medical applications, industrial and commercial applications, and there is room for unconventional ones. For future, we expect that the parallel processing especially on GPUs will help more time-critical applications to emerge. The evolution of the transform will keep going. As the transform has had a fruitful history, it has a good chance of a bright future for decades to come.

**Allam S. Hassanein** obtained his B.Sc. degree in Computer and Electronics Engineering from the Higher Thebes Institute for Engineering in Egypt. Then, received his M.Sc. degree in Computer Science and Engineering from Cairo University in 2011. He had worked as a teaching assistant in the Higher Thebes Institute for Engineering, then as a lecturer at the College of Computer, Qassim University, KSA. Currently, he is a research associate at the Informatics Department, the Electronics Research Institute, Egypt. He was the CO-PI in a two year-research project for developing an assisting system for the blind. The project was funded from Tibah University, KSA. At present, he is a PhD student at Egypt-Japan University for Science and Technology on MOHE Grant for PhD Studies. He published in international journals and conferences. His recent research interests include computer vision, image processing, artificial intelligence, pattern recognition, sign language recognition, and humanoid robots.

**Sherien Mohamed** obtained her B.Sc. degree in Computer Engineering from Cairo University. She received her M.Sc. degree from the same department in 2011. She worked as a teaching assistant in MSA University, Egypt from 2003 to 2013. Currently, she is a research associate at the Informatics Department, the Electronics Research Institute, Egypt. She published in international journals and conferences. Her recent research interests include computer vision, robotics, and human activity identification.

**Mohamed Sameer** received his B.Sc. in Computers and Systems from the Faculty of Engineering, Al-Azhar University. He obtained his M.Sc. degree in Computer Engineering from the Faculty of Engineering, Cairo University in 2012. Currently, he is a research associate at the Informatics Department, the Electronics Research Institute, Egypt. He published in international journals and conferences. His recent research interests include image processing, computer vision and intelligent systems.

**Mohammad Ehab Ragab** is a Researcher at the Informatics Department, the Electronics Research Institute in Egypt. He received the B.Sc., M.Sc., degrees from Ain Shams University, and Cairo University respectively. He obtained his Ph.D. from the Chinese University of Hong Kong (CUHK) in 2008. During his undergraduate studied, he received the "Simplified Scientific Writing Award" from the Academy of Scientific Research and Technology. He was awarded the "IDB Merit Scholarship" for conducting his Ph.D. studies. He taught courses in some distinguished Egyptian universities, and published in international journals and conferences. His research interests include: Computer vision, Robotics and Image Processing.